%% file: main.tex
\documentclass[11pt,a4paper]{article}
\usepackage{graphicx}
\graphicspath{ {./images/} }
\usepackage{multirow}
\usepackage{times,latexsym}
\usepackage{url}
\usepackage[T1]{fontenc}
\usepackage{fdsymbol}

\usepackage{acl}
\usepackage{xspace}

\usepackage{hyperref}

\newcommand{\gptfrozen}[0]{GPT-2{\tiny{frozen}}}
\newcommand{\gptlmft}[0]{GPT-2{\tiny{HLC}}\xspace}
\newcommand{\name}[0]{HaRT\xspace}
\newcommand{\hulm}{\textsc{HuLM}\xspace}
\newcommand{\hulmdataset}[0]{HuLM-Corpus\xspace}

\title{Human Language Modeling}

\author{Nikita Soni, Matthew Matero, \\ {\bf Niranjan Balasubramanian, \and H. Andrew Schwartz} \\
Department of Computer Science, Stony Brook University \\
\texttt{\{nisoni, mmatero, niranjan, has\}@cs.stonybrook.edu}}

\date{}

\begin{document}
\maketitle
\begin{abstract}
Natural language is generated by people, yet traditional language modeling views words or documents as if generated independently. 
Here, we propose \textit{human language modeling} (HuLM), a hierarchical extension to the language modeling problem whereby a human-level exists to connect sequences of documents (e.g. social media messages) and capture the notion that human language is moderated by changing human states. 
We introduce, \name, a large-scale transformer model for the \hulm task, pre-trained on approximately 100,000 social media users, and demonstrate its effectiveness in terms of both language modeling (perplexity) for social media and fine-tuning for 4 downstream tasks spanning document- and user-levels: stance detection, sentiment classification, age estimation, and personality assessment.\footnote{Code and pre-trained models available at: \href{https://github.com/humanlab/HaRT}{https://github.com/humanlab/HaRT}.} Results on all tasks meet or surpass the current state-of-the-art.
\end{abstract}

\input{intro}

\input{related}

\input{hulm}
\input{eval}

\input{conclusions}
\bibliography{main}
\bibliographystyle{acl_natbib}

\input{appendix}

\end{document}

%% file: intro.tex
\section{Introduction}
Language use, like any human behavior, is moderated by underlying human states of being~\cite{mehl2003sounds,fleeson2001toward}. Indeed, different ways of incorporating human information into NLP models have recently been shown to improve accuracy on many NLP tasks~\cite{hovy-2015-demographic,lynn2017human, huang2019neural,hovy-yang-2021-importance}.
At the same time, while language modeling has proven itself fundamental to NLP, it is typically absent the notion of a human producing the natural language.

From a statistical modeling perspective, this absence of human state can be seen as an instance of the \textit{ecological fallacy} -- the treatment of multiple observations (i.e. text sequences) from the same source (i.e. human) as independent~\cite{piantadosi1988ecological,steel1996analysing}. 

To address this, we introduce the task of \textit{human language modeling} (\hulm), which induces dependence among text sequences via the notion of a human state in which the text was generated. 
In particular, we formulate \hulm as the task of estimating the probability of a sequence of tokens, $w_{t,1:i}$, while conditioning on a higher order state ($\mathbf{U}_{1:t-1}$) derived from the tokens of other documents written by the same individual. Its key objective is:  
\vspace*{-6pt}
\begin{equation*}\vspace*{-6pt}
Pr(w_{t,i} | w_{t,1:i-1}, \mathbf{U}_{1:t-1})
\end{equation*}
where $t$ indexes a particular sequence of temporally ordered utterances (e.g. a document or social media post), and $\mathbf{U}_{1:t-1}$ represents the human state just before the current sequence, $t$. 
In one extreme, $\mathbf{U}_{1:t-1}$ could model all previous tokens in all previous documents by the person. In the opposite extreme, $\mathbf{U}_{1:t-1}$ can be the same for all users and for values of $t$ reducing to standard language modeling: $Pr(w_{i} | w_{1:i-1})$.\footnote{See section \ref{sect:hulm} for a full \hulm definition.}
Thus, \hulm-based models without history can be used where traditional LMs are applied (and may even perform better).

\hulm brings together ideas from human factor inclusion/adaptation~\cite{hovy-2015-demographic,lynn2017human, hovy-yang-2021-importance} and personalized modeling~\cite{king-cook-2020-evaluating,jaech-ostendorf-2018-personalized} into the framework of large pre-trained language models.
Compared to traditional language modeling, \hulm offers several technical advantages. 
First, the human state serves as a higher order structure that induces dependence between the text sequences of the same person/  thus posing a language modeling problem that is a more faithful treatment of human-generated natural language. 
Second, conditioning on prior texts of an individual can be seen as an implicit integration of text-derived human factors without having to explicitly model the identity of the individual. This enables fine-tuning of such a model to many downstream tasks. Third, using the temporally ordered prior texts for human contexts can be seen as a way to track the dynamic nature of human states (e.g. emotions, daily activities) and be combined to yield more stable personality traits (e.g, extraversion, openness).

To build a language model that effectively addresses the \hulm task, we develop \name, a human-aware recurrent transformer. \name is built using a new user-state based attention layer, that connects standard word sequence transformer layers in order to incorporate the human context. The recurrent user state allows \name to effectively model long contexts necessary to handle all the previous messages written by an individual.
We train \name on the \hulm task defined over a large collection of social media texts spanning 100K users and apply it (fine-tuning) on 2 downstream message-level tasks: stance detection~\cite{StanceSemEval2016}, and sentiment analysis~\cite{nakov2013task} as well as 2 human-level tasks: age estimation and personality assessment~\cite{schwartz2013personality}.

\paragraph{Contributions.}
Our contributions are three-fold: (1) We introduce the task of human language modeling (\hulm), providing a mathematical definition and relation to traditional language modeling; (2) We propose \name, a novel transformer-based model for performing \hulm and capable of being fine-tuned to specific tasks including user-level tasks for which traditoinal language models cannot be applied without architectural alterations; (3) We evaluate \name, demonstrating state-of-the art performance on five tasks: social media language modeling (perplexity), two document-level tasks (sentiment analysis and stance detection), and two user-level tasks (personality--openness assessment, and age estimation).

%% file: related.tex
\section{Related Work}
Recent advances in language model pre-training have led to learned representation of text. Pretraining methods have been designed with different training objectives, including masked language modeling \cite{devlin2019bert} and permutation-based auto-regressive language modeling \cite{yang2019xlnet}. These have contributed in building deep {\textit{autoencoding}} architectures, allowing the same pre-trained model to successfully tackle a broad set of NLP tasks.
While pre-training over large collections of text helps models acquire many forms of linguistic and world knowledge\cite{Petroni2019LanguageMA, Jiang2020HowCW, Rogers2020API}, they are still devoid of the information about the text creator. 

Recently, it has been suggested that the NLP community 
address the social and human factors to get closer to the goal of human-like language understanding~\cite{hovy-yang-2021-importance}.
This call builds on a series of studies suggesting that integrating the human context into natural language processing approaches leads to greater accuracy across many applications in providing personalized information access~\cite{dou2007large, teevan2005personalizing} and recommendations \cite{guy2009personalized, li2010contextual, de2012chatter}. 
The idea of contextualizing language with extra linguistic information has been the basis for multiple models:
\citet{hovy-2015-demographic} learn age- and gender-specific word embeddings, leading to significant improvements for three text classification tasks. \citet{lynn2017human} proposed a domain adaptaion-inspired method for composing user-level, extra-linguistic information with message level features, leading to improvements for multiple text classification tasks. \citet{welch-etal-2020-compositional} propose a new form
of personalized word embeddings that use
demographic-specific word representations. 


In addition to addressing to social and human factors, recent work has also focused on \textit{personalized} language models ~\cite{king-cook-2020-evaluating,jaech-ostendorf-2018-personalized} learning author representations \cite{delasalles2019learning} and  personalized word embeddings \cite{lin2017personalized} pointing out the importance of personalized semantics in understanding language. \citet{welch-etal-2020-exploring} explore personalized versus generic word representations showing the benefits of both combined. While these models are trained for singular user, \citet{mireshghallah2021useridentifier} trains a single shared model for all users for personalized sentiment analysis. However, the approach is not scalable as it is still user specific and expects a unique user identifier.
 

While not the primary goal, human language modeling may yield effective approaches to extend the context during language modeling. 
Thus, an aspect of this work can be seen as part of the recent pursuit of sequence models that capture dependencies beyond a fixed context length \cite{Dai2018TransformerXLLM, beltagy2020longformer}. 
For example, \citet{keskar2019ctrl} and \citet{dathathri2019plug} propose controllable language generation using one or more attribute classifiers or control codes. 
\citet{guu2020realm}  propose augmented language model pretraining with a latent knowledge retriever which
allows the model to retrieve and attend over documents from a large corpus. 
These models extend context limits, but they do not model the higher order structure capturing a notion of the common source of documents i.e., the author.
On the other hand, \citet{yoshida2020adding} fits a hierarchical model extension to language modeling by adding recurrence to a pretrained language model. 
This idea forms a basis for our proposed \hulm architecture, \name, but \citeauthor{yoshida2020adding} do not exploit the inherent higher order structure (i.e. the model was not used for \hulm). 

%% file: hulm.tex
\section{Human Language Modeling (\hulm)}
\label{sect:hulm}

Our goal is to re-formulate the language modeling task into one that directly enables a higher-order dependence structure that represents a human generating the language. 
 
Language modeling formulations pose probabilistic questions over text represented as sequences of tokens. The main goal is to model the probability of observing a given token sequence in the language as a whole. In particular language models (LMs) estimate the joint probability of the tokens in the string, defined in terms of the probabilities of each token in the sequence conditioned on the previous tokens.\footnote{Traditional LMs provide estimates of the conditional probabilities often relying on further simplifying assumptions (e.g. Markovian assumptions to handle long sequences.).} Given a string $\mathbf{W}\in \mathbb{L}$, a sequence of $n$ tokens $\langle w_1, w_2, \cdots, w_n \rangle$, the probability of observing the string $\mathbf{W}$ in the language $\mathbf{L}$ is computed as:
\begin{equation}
\label{eq:lm}
Pr(\mathbf{W}) = \prod_{i=1}^n Pr(w_{i} |w_{1:i-1})     
\end{equation}

We pose the \emph{human language modeling} problem (HuLM), where the goal is to model the probabilities of observing a sequence from the language as generated by a specific person. An initial idea might be to pose this task as conditioning the probability of a string, $w_i$ on a static representation of the person (or user, $\mathbf{U}_{static}$): 
\begin{equation}
\label{eq:static-user-lm}
Pr(\mathbf{W}|\mathbf{U}_{static}) = \prod_{i=1}^n Pr(w_{i} | w_{1:i-1}, \mathbf{U}_{static} )
\end{equation}
This addresses the first of the two goals we presented in the introduction, namely avoiding the \textit{ecological fallacy} of assuming sequences from the same person are independent. However, it does not respect the idea that people vary in mood and can change. More precisely, human behaviors (language use) are influenced by dynamic human states of being~\cite{fleeson2001toward,mehl2003sounds}. Thus, we pose HuLM with a more general formulation that enables the idea of a dynamic representation of humans, the user state $\mathbf{U}_{t}$\footnote{We define $\mathbf{U}_{t}$ as the state \textit{after} the sequence, $\mathbf{W}_t$. Thus, only $\mathbf{U}_{t-1}$ is accessible as given when estimating Pr($\mathbf{W}_t$) conditioned on the user state.}: 
\begin{equation}
\label{eq:word_given_userstate}
Pr(\mathbf{W}_t|\mathbf{U}_{t-1}) = \prod_{i=1}^n Pr(w_{t,i} | w_{t,1:i-1}, \mathbf{U}_{1:t-1})
\end{equation} 
where $t$ indexes a particular sequence of temporally ordered utterances (e.g. a document, or set of social media message). 
While $w_{t,i}$ is drawn from a multinomial distribution, $\mathbf{U}_{1:t-1}$ can be from any discrete or continuous multivariate distribution. 

In one extreme, $\mathbf{U}_{1:t-1}$ could model all previous tokens in all previous documents by one person.  In the opposite extreme, $\mathbf{U}_{1:t-1}$ can be the same for all values of $t$, giving a static representation for a user (equivalent to \autoref{eq:static-user-lm}) or even static across users which reduces to a standard language modeling version (equivalent to \autoref{eq:lm}). Still, modeling a user via their previous documents provides a seamless way to integrate the user information into language models -- the only change is that the models will now have to incorporate more text when they are making predictions.
Note that this problem formulation does not directly require explicit modeling of the identity of a user. This makes it easier to handle new users in downstream tasks and test instances, or creating models that can be further fine-tuned to both document- and user-level tasks.

\paragraph{HuLM in Practice.}
Like traditional langauge models, there are two steps to applying HuLM based models to most tasks and applications: pre-training and fine-tuning. 
During pre-training, the model is trained on unlabeled data over Human Language Modeling (HuLM) pre-training task above. 
For finetuning, a HuLM based model is first initialized with
the pre-trained parameters, and all of the parameters are fine-tuned using labeled data from the downstream tasks. Each downstream task has separate fine-tuned models, even though they are initialized with the same pre-trained parameters.

\section{Human-aware Recurrent Transformer}

This section introduces, \name, a human-aware recurrent transformer that trains on the human language modeling (\hulm) formulation. 

\name is designed to produce human-aware contextual representations of text at multiple levels. \name's design is motivated by two goals: (i) We want to support  hierarchical modeling, i.e., to hierarchically represent the set of all-messages written by a user and at the same time have human-aware contextual word representations. This implicitly entails modeling large context size. For example, GPT-2 \cite{radford2019language} uses a context size of 1024 tokens, whereas our estimate of the average context size for a Twitter user is more than 12000 tokens. (ii) To support user-level tasks (e.g. personality assessment \cite{lynn-etal-2020-hierarchical}), we need representations of the entire set of messages written by a user capturing the inherent human states that broadly encompasses the user representation.

\name addresses the hierarchical language modeling issue by processing all messages written by a user in a temporally ordered sequence of blocks. It  uses a recurrence structure to summarize information in each block  into a user state vector, which is then used to inform the attention between tokens in the subsequent block. 
For human-level tasks the aggregate of user states can be used as the representation of the entire context for the user.

The idea of adding recurrence to pre-trained transformers builds on \citet{yoshida2020adding}'s method for handling long contexts. 
However, the main difference is that \name models the input data (language) in the context of its source (user) along with inter-document context, thus enabling a higher order structure representing human context.

\subsection{\name Architecture}

\begin{figure}
    \centering
    \includegraphics[width=0.50\textwidth]{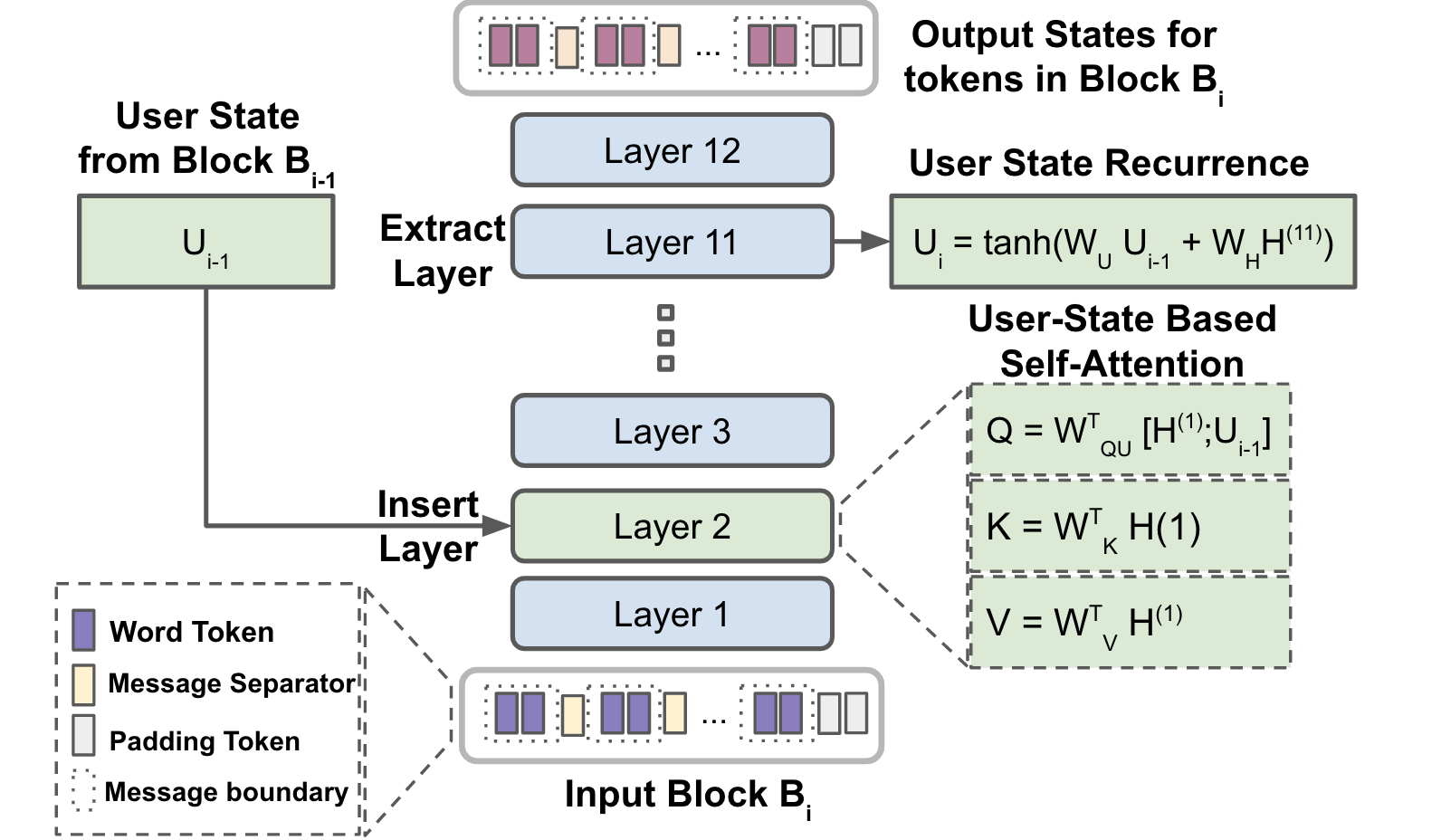}
    \caption{\name architecture: \name processes a user's messages in blocks. It produces contextualized representations of messages in each block conditioning on a recurrently computed user state. The user state is inserted into an earlier layer (layer 2) to inform the self-attention computation via a modified query transform. The previous user state is then recurrently updated using the output of a later layer (layer 11). }
   
    \label{model_fig}
\end{figure}
Figure~\ref{model_fig} shows the overall architecture for \name. It consists of a one modified transformer layer with a user-state based self-attention mechanism over more token-level standard self-attention based transformer layers from a pre-trained transformer (GPT-2). 

\noindent{\textbf{Inputs and Outputs}} Each input instance to \name consists of a temporally ordered sequence of messages (by message created time) from a given user $a$, $\mathcal{M}_a = \langle M_1, \cdots, M_n \rangle$. We segment these messages into fixed sized blocks, $\mathcal{B}_a = \langle B_1, \cdots, B_k \rangle$. We sequentially fit messages into blocks, separating messages using a newly introduced special token ${<|insep|>}$. If the number of tokens in a block falls short of the block size, we fill it with padded tokens. $k$ is a hyperparameter during training used to cap the maximum number of blocks controlling the amount/size of user history that is fed to the model. If the messages for a user fill less than $k$ blocks, we pad the rest to maintain the same size for each instance.

For each block $B_i$, \name outputs (i) contextualized representations of the tokens within the block conditioned on the previous user state ($U_{i-1}$), and (ii) an updated representation of the user state, $U_i$, which now also includes the information from the current block $B_i$. 
We use the representation of the last non-pad token of a message as its representation for message-level tasks, and use the average of the user-states from all the blocks of a user as that user's representation for user-level tasks.

\noindent{\textbf{User-State based Self-Attention}} \name constructs a user-state representation vector by combining information from each block in a recurrent manner. After processing the inputs in a given block $B_i$, \name extends the previous user state $U_{i-1}$ with information from current block $B_i$ using the output representations $H^{(E)}$ from one of the later layers (we denote as the extract layer $L_E$). The recurrence for the new user state $U_i$ is:

\begin{equation}
U_{i} = tanh(W_{U} U_{i-1} + W_H H^{(E)})
\end{equation}

The user state for the first block $U_0$ is initialized with the average of the (pretrained GPT-2) layer 11 outputs for words from the messages of more than 500 users (of the train set) computed using \citet{schwartz2017dlatk}. 

To produce the user-state conditioned contextual representations at a given layer, \name uses a modified self-attention procedure to one of the earlier layers, which we denote as the insert layer ($L_{IN}$).
The idea is to create a new query transform which includes the user-state vector, so that the attention between tokens is informed by the context of the previous messages written by the user. To this end, we take input hidden states to this insert layer $H^{IN-1}_i$, concatenate it with the user-state vector from the previous block $U_{i-1}$ 
and then apply a linear transformation (using $W_q$) to obtain the query vectors ($Q^{IN}_i$) for the self-attention computation. 

\begin{equation}
\it{Q^{IN}_{i}} = \it{W_q^T}[\it{H^{(IN-1)}_i};\it{U_{i-1}}] 
\end{equation}

The key, value transforms and the rest of the self-attention computation and further processing in the transformer to produce the output representations from the layer, all remain the same as in the original GPT-2 model. 

\noindent{\textbf{Implementation Choices}} There are multiple alternatives for a \name implementation including how to construct the user state, where and how to inject user state information. In our preliminary experiments we experimented with different extract layers but found that constructing user state from the penultimate layer (Layer 11) and injecting the user state in a single earlier layer (Layer 2 used by \citet{yoshida2020adding})  to modify the query transformation was the most effective empirically.

\subsection{Pre-training \name}
\label{ulm_task}

\name is pre-trained using the \hulm task in an autoregressive manner. 

The \hulm task as defined in Equation~\ref{eq:word_given_userstate} asks to predict a token that appears in a token sequence (i.e. a user's social media message) given the previous tokens in the sequence while also conditioning on previous user states. We turn this task into a pre-training objective defined over block segmented token sequences from a user. 
For each block of a given user, the task is to predict each token in the block while conditioning on (i) the previous tokens within the current block which are directly available as input, and also (ii) the tokens from the previous blocks that are available to \name through the recurrent user state. 
Formally, the pre-training objective is to maximize:
\begin{equation}
\prod_{a \in \textup{Users}}
\prod_{t = 1}^{|\mathcal{B}_a|} \prod_{i=1}^{|B_{t}^{(a)}|} Pr(w_{t,i} | w_{t,1:i-1}, B_{1:t-1}^{(a)})
\end{equation}
where, $w_{t,i}$ is the $i^{th}$ token in the $t^{th}$ block ($B_{t}^{(a)}$) for user $a$.

\textbf{Pre-training data} For the pre-training corpus we combine a subset of the Facebook posts dataset from \citet{Park2015AutomaticPA}, 
~a subset of the County Tweet Lexical Bank \citep{giorgi-etal-2018-remarkable} appended with newer 2019 and 2020 tweets, in total spanning 2009 through 2020.  
We filter the datasets to only include tweets marked as English from users who have at least 50 total posts and at least 1000 words in total, ensuring moderate language history for each user. 
The resulting dataset consists of just over 100,000 unique users, which we split into a train dataset consisting of messages from 96,000 users, a development dataset that consists of messages from 2000 users that were not part of the training set (unseen) and new messages from 2500 users seen in the training set, and a test set of messages from a separate set of 2000 unseen users that are neither in training or the development set. 

We refer to this as the \hulmdataset (HLC).

\subsection{Fine-tuning \name}
\label{FT_hulm}

In the tradition of transformers for traditional language modeling, {\name} shares the same architecture for both pre-training and fine-tuning except for the output layers. It has a unified architecture across different downstream tasks. For finetuning, \name is first initialized with
the pre-trained parameters, and all of the parameters are fine-tuned using labeled data from the downstream tasks. Each downstream task has separate fine-tuned models, even though they are initialized with the same pre-trained parameters.
Apart from using the labeled data from the downstream tasks, we also use the historical messages (when available) from the respective users to replicate the format of pre-training inputs and to benefit from the knowledge of the user. 

%% file: eval.tex
\section{Evaluation: Human Language Modeling}
\label{eval_lm}

We seek to compare \name with a standard language model that is exposed to the same data but without modeling the notion of a user. Thus, we compare \name's human language modeling performance to the model it was based, GPT-2. For calibration we report performance on GPT-2's original pre-trained version (\gptfrozen), and a version of the LM that was fine-tuned on the \hulmdataset (\gptlmft). 

We train and evaluate the models using the train and test splits of the \hulmdataset described in Section~\ref{ulm_task}. For hyperparameter search, we use  the full development set of both seen and unseen users. 
Each training instance for \name is capped to 8-blocks of 1024-tokens each. Following previous work fine-tuning transformer language models for social media~\cite{v-ganesan-etal-2021-empirical}, GPT-2 was trained over individual messages. 
We train both for five epochs and set the learning rate, batch size, and stopping patience based on the development set (see Appendix \ref{expmt_settings}). 
For \name, we initialize all GPT-2 self-attention layers with the corresponding weights in the pre-trained GPT-2. 
The user-state based self-attention layer weights (query, key, and value) are normal initialized with 0 mean and 0.02 standard deviation.

\paragraph{Perplexity}
Table~\ref{tab:ppl} reports the perplexity of all three models on the messages from the unseen users of the development split and the entire test split of \hulmdataset. The frozen pre-trained GPT-2 (\gptfrozen) fares poorly to the domain mismatch while the fine-tuned version (\gptlmft) fares much better. However, the human language model \name achieves the best performance by a large margin, with a significant reduction in perplexity by more than 46\% on the test set relative to \gptlmft ($p < .001$).\footnote{In addition to this improvement for unseen users, we also see similar relative benefits when tested on instances from seen users which we report in Appendix \ref{eval:seen_ppl}.}

\begin{table}[t]
\begin{center}
\begin{tabular}{|l|p{1.6cm}|p{1.9cm}|p{1.6cm}|p{1.9cm}|}
\hline \bf Model  &  Dev ($ppl$)  &  Test  ($ppl$) \\ 
\hline
\gptfrozen & 112.82  & 116.35 \\
\gptlmft & 47.61 &  48.51  \\
\name  & \bf{27.49*} &  \bf{26.11*} \\
\hline
\end{tabular}
\end{center}
\caption{\label{tab:ppl} Comparing \name as a language model to \gptfrozen, the frozen pre-trained GPT-2 and \gptlmft, the GPT-2 model fine-tuned on the \hulmdataset. \name shows large gains with a substantial reduction in perplexity compared to both versions of GPT-2. Bold font indicates best in column and * indicates statistical significance $p < .05$ via permutation test w.r.t \gptlmft}
\end{table}

\paragraph{Effect of History Size.}

\begin{figure}[tb]
    \centering
    \includegraphics[width=0.5\textwidth]{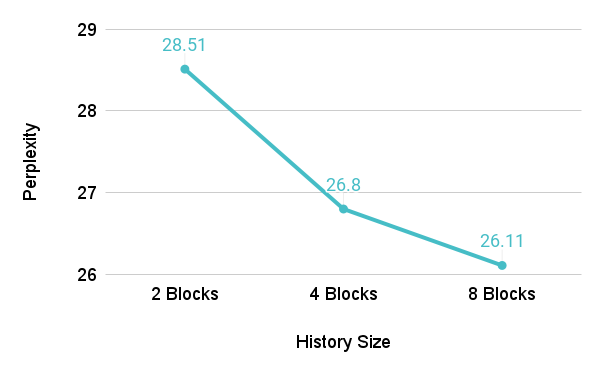}
    \caption{\label{fig:ppl_hist} . Perplexity scores, on test sets 
    as a function of history size (number of blocks) used when training \name. Each block consists of 1024 tokens. Adding more history improves language modeling performance with big reduction going from 2 to 4 blocks and a smaller reduction from 4 to 8 blocks.}
    
\end{figure}

We further analyze the effect of history size by varying the amount of language, in terms of blocks, used per user. Figure~\ref{fig:ppl_hist} shows that adding more history in general helps, with a big reduction in perplexity going from 2 to 4 blocks and a further reduction going from 4 to 8 blocks. Adding more context can induce a need to effectively balance likelihood of finding more important signals against the increasing chances of it drowning in less important information.

\section{Evaluation: Fine-tuning for Downstream Tasks}
Here, we evaluate the utility of fine-tuning \name for document- and user-level tasks.
Just as standard transformer language models are fine-tuned for tasks, we take our pre-trained {\name} model and fine-tune it for stance detection, sentiment classification, age estimation, and personality (openness) assessment tasks. 
For both sets of tasks we compare fine-tuning the \gptlmft as a non-user-based LM baseline and also report previously published results from other task specific models, most of which employ historical context for respective tasks. All hyperparameter settings and training details for the \gptlmft and \name models for each task are listed in Appendix \ref{expmt_settings}.

\begin{table}[t]
\begin{center}
\begin{tabular}{|p{1.5cm}|p{1cm}|p{1cm}|p{1cm}|p{1.5cm}|}
\hline \bf Model & \it{Age ($r$)} & \it{OPE ($r_{dis}$)} & \it{Stance ($F1$)} & \it{Sentiment ($F1$)}  \\ 
\hline

\gptlmft  & 0.839 & 0.521  & 68.60 & 76.75 \\
\name  & \bf{0.868*} & \bf{0.619*} & \bf{71.10*} & \bf{78.25*}  \\
\hline
\end{tabular}
\caption{\label{tab:all_ft} We fine-tune \name and \gptlmft (GPT-2 fine-tuned for LM on the same data) for 4 downstream tasks: Age, Openness (OPE), Stance, and Sentiment, and find \name to perform better on all 4 tasks. For age and openness, we fine-tune \name only for the recurrence module, and fine-tune only the last 2 layers of \gptlmft. For stance and sentiment, we fine-tune full models. Results are reported in pearson r for Age, disattenuated pearson r for OPE and weighted F1 for Stance/Sentiment. Bold indicates best in column and * indicates statistical significance $p < .05$ via permtuation test.   
}
\label{tab:all_ft}
\end{center}
\end{table}

\subsection{Document-Level Tasks}
\label{ft_tasks}
\label{doc_ft_tasks}

We consider two document-level tasks that require models to read an input document (message) written by a user and output a label (stance of the user towards a topic or the sentiment expressed in the text). To fine-tune \name on these tasks, with each document we collect and attach previous messages written by the same users, represented using the procedure we outlined in Section~\ref{FT_hulm}. Thus, \name processes this input to produce message- and human-contextualized token-level representations. We represent the document by its last non-padded token representation and feed it to classification layer with a prior layer norm for predicting the output label. 
\gptlmft, without hierarchical structure, only uses the input document to make predictions. 
We fine-tune all parameters of \name and \gptlmft, as well as the classification layer weights using the standard cross-entropy loss (calculated only over the last non-padded token of the target (labeled) messages).

\paragraph{Stance Detection.}
For stance detection we use the SemEval2016 dataset \cite{StanceSemEval2016}, which contains tweets annotated as being in favor of, against, or neutral toward one of five targets: atheism, climate change as a real concern, feminism, Hillary Clinton, and legalization of abortion. This data only includes labeled tweets from users and not any history, so we use the extended dataset from \citet{lynn-etal-2019-tweet} and preserve the train/dev/test split of the same. To maintain (message created time) temporal accuracy in our autoregressive model, we only used the part of the extended dataset (history) that consists of messages posted earlier than the labeled messages. 
\paragraph{Sentiment Analysis.}
We use message-level sentiment annotations indicating positive, negative, and neutral categories from the SemEval-2013 dataset \cite{nakov2013task}. As with stance, we use a part of the extended dataset from \citet{lynn-etal-2019-tweet} to get associated message history, and preserve the train/dev/test split of the same.

\begin{table}[t]
\begin{center}
\begin{tabular}{|l|p{1cm}|p{1.5cm}|}
\hline \bf{Model} &
\it{Stance (F1)} &
\it{Sentiment (F1)} \\ 
 
\hline
MFC & 54.2 & 28.0\\
\citet{lynn-etal-2019-tweet} &65.9 & 69.5\\
MeLT & 66.6 & 63.0 \\
BERTweet & 68.8 & 77.9 \\

\name & \bf{71.1*} & \bf{78.3*} \\
\hline
\end{tabular}
\caption{\label{tab:sota-doc-ft} We compare \name's performance on document level downstream tasks: Stance and Sentiment, against state of the art results. We also fine-tuned pre-trained GPT-2, BERTweet \cite{nguyen2020bertweet}, and MeLT~\cite{matero-etal-2021-melt-message} on both tasks for baselines. \name performs the best in both tasks with a substantial gain. Results are reported in weighted F1. Bold indicates best in column and * indicates statistical significance $p < .05$ w.r.t BERTweet via permutation test.
}
\end{center}
\end{table}

\subsection{User-Level Tasks}
\label{user_ft_tasks}
We evaluate \name for age estimation and personality (openness) assessment, social scientific tasks which require producing outcomes at the user-level. We use a subset of the data from consenting users of Facebook who shared their Facebook posts along with demographic and personality scores~\cite{Kosinski5802,Park2015AutomaticPA}.

For these user-level tasks we can leverage the recurrent user states in \name to produce a representation of the user. We represent the input as described in Section \ref{FT_hulm}, and use the average of the user-states vectors from the non-padded blocks of each user and layer norm it to make predictions using a linear classifying layer to predict 1 label (regression task). We use only 4 blocks of history when training to fine-tune.

For \gptlmft, since it can't directly handle all of the users text in one go, we replicate the user label for each message of the respective users and train the model to predict the label for each message using the last non-padded token of the message. To make the final prediction, we average the predictions across all messages from respective users and calculate the performance metric using this average as in~\cite{v-ganesan-etal-2021-empirical}.

For these user level tasks that require aggregate information, for both models, fine-tuning the entire set of parameters was worse than fine-tuning fewer layers. For \gptlmft fine-tuning only the last two layers gave the best performance. For \name fine-tuning only the recurrence module gave the best performance on development sets. We report results with these best dev settings. We use the mean squared error (MSE) as the training loss.

\paragraph{Age Estimation}
Similar to the pre-training data, we filtered the above dataset for English language instances and included only the users with a minimum of 50 posts and a minimum of 1000 words. Age was self-reported and limited to those 65 years or younger. This resulted in a dataset of 56,930 users in train, 1836 users in dev, and 4438 users in test which was a subset of the test set (5000 users) from \citet{Park2015AutomaticPA}. We evaluate on both the test sets and report Pearson correlation ($r$) metric on the latter for comparison purposes.
We include results with the filtered data in Appendix (Table \ref{tab:user_ft_filtered}).

\paragraph{Personality Assessment.}
We evaluate on the assessment of openness based on language (one’s tendency to be open to new ideas) \cite{schwartz2013personality}.  To allow for direct comparisons, we use the same test set (n=1,943) as \citet{lynn-etal-2020-hierarchical} and use a  subset of their training set (66,764 users) of which 10\% were sampled as dev set, and report disattenuated pearson correlation ($r_{dis}$) to account for questionnaire reliability \citet{lynn-etal-2018-clpsych}. As with age estimation, we report results with the filtered dataset in Appendix (Table \ref{tab:user_ft_filtered}).

\begin{table}[t]
\begin{center}
\begin{tabular}{|p{3.6cm}|l|l|}
\hline \bf{Model} &
\it{Age ($r$)} & 
\it{OPE ($r_{dis}$)} \\ 
 
\hline
\citet{v-ganesan-etal-2021-empirical} & 0.795 & 0.511  \\
\citet{sap-etal-2014-developing} & 0.831 & -  \\
\citet{lynn-etal-2020-hierarchical} & - & \textbf{0.626}\\
\name & \textbf{0.868*} & \textbf{0.619} \\
\hline  
\end{tabular}
\caption{\label{tab:sota-user-ft} Comparison of \name's performance on user level downstream tasks: Age and Openness (OPE), against state of the art results. \citet{v-ganesan-etal-2021-empirical} use lesser number of users (10000) in training. 
Results are reported in pearson r for Age and disattenuated pearson r for OPE. Bold indicates best in column and * indicates statistical significance between \name and \cite{sap-etal-2014-developing} ($p < .05$) using a bootstrap sampling test. We also find no statistical difference between \name and \cite{lynn-etal-2020-hierarchical} ($p=.35$).
}
\end{center}
\end{table}

\subsection{Results}
\label{results}

Table~\ref{tab:all_ft} summarizes the performance of \name against the baseline of fine-tuning a non-human-aware language model, \gptlmft. We see that \name yields substantial gains over \gptlmft across both user-level and document-level tasks, demonstrating clear benefits in all settings.

\noindent{\textbf{Document-Level Tasks}}
Table~\ref{tab:sota-doc-ft} compares \name with task-specific baselines for stance and sentiment detection including (i) \citet{lynn-etal-2020-hierarchical} which used historical contexts to incorporate both explicit and text-derived latent human factors, (ii) MeLT~\cite{matero-etal-2021-melt-message} which used a superset of the same historical contexts used here but for message-level language modeling, and (iii) BERTweet~\cite{nguyen2020bertweet} which uses a large collection  of tweets to pretrain an autoencoder that is then fine-tuned for target tasks.
Sentiment results are weighted F1 scores over the three sentiment categories. 
Stance results are an average of weighted F1 scored over five different topics from respective topic-specific fine-tuned models. \name outperforms all models demonstrating the substantial benefits of human language modeling for these document-level downstream tasks.

\noindent{\textbf{User-Level Tasks}} Table~\ref{tab:sota-user-ft} compares \name with task-specific baselines for Age and Openness tasks that use the superset of the same data used by \name. For Age, \name outperforms all baselines including a strong non-neural lexica based predictor~\cite{sap-etal-2014-developing}, and a RoBERTa-based system that uses carefully chosen frozen embeddings~\cite{v-ganesan-etal-2021-empirical}. For Openness, \name is better than the frozen RoBERTa \cite{liu2019roberta} embeddings and is comparable to \citet{lynn-etal-2020-hierarchical}'s hierarchical attention model. These results also suggest the potential of \name's user states as a representation for user-level tasks.

\subsection{No Historical Context.}
\name can also be used anywhere a typical transformer language model is used by simply not feeding any historical context. 
Here, we seek to use our pre-trained \name as a language model that is fine-tuned to the messages  (for the respective tasks) without any historical context. 
Table~\ref{tab:data_wo_history} compares the performances of \name and \gptlmft for the two document-level downstream tasks Stance, and Sentiment. For a fair comparison, we use the same data inputs for both the pre-trained models which consists of only the labeled messages and no historical context. We evaluate in 2 ways: 1) freezing the model and training only the classification layer using the outputs from the penultimate transformer layer, and 2) fine-tuning all model parameters along with a classification head with a layer norm prior to it. \name is at par or better with \gptlmft for both frozen and fine-tuned versions, showing that it can provide gains even when historical context is unavailable. 
Hyperparameters settings are described in Appendix \ref{expmt_settings}.

\begin{table}[t]
\begin{center}
\begin{tabular}{|p{2.4cm}|p{1.5cm}|p{1cm}|}
\hline \bf Model  & \it{Sentiment ($F1$)} & \it{Stance ($F1$)}  \\ 
\hline

\gptlmft \tiny{frozen}  & 62.7 & 57.7 \\
\name \tiny{nohist, frozen}  & 62.7 & 58.6 \\

\gptlmft  &  76.8 & 68.6  \\

\name \tiny{nohist} & \bf{77.7*} & \bf{70.8*} \\
\hline
\end{tabular}
\end{center}
\caption{\label{tab:data_wo_history} Results with experiments on Stance and Sentiment downstream tasks using only the labeled instances and no history. We compare \name with \gptlmft by training only the classification head (\textit{frozen}) and additionally, by fine-tuning the models. Bold indicates best in column and * indicates statistical significance $p < .05$ via permutation test w.r.t \gptlmft. Results are reported in weighted F1.}
\label{tab:ablation} 

\end{table}

\subsection{Ablation Studies}
In this section, we perform ablation experiments on {\name} to better understand their relative importance and report the results in Table~\ref{tab:ablations}.

\begin{table}[t]
\begin{center}
\begin{tabular}{|p{2.4cm}|p{1.5cm}|p{1cm}|}
\hline \bf Model  & \it{Sentiment ($F1$)} & \it{Stance ($F1$)} \\ 
\hline
\name \tiny{NOT PT} & 63.47 & 66.26 \\

\name \tiny{W/O RECUR} &  77.04 & 68.73 \\

\name & \bf{78.25*} & \bf{71.10*} \\
\hline
\end{tabular}
\end{center}
\caption{\label{tab:ablations} Results with the ablation experiments on Stance and Sentiment downstream tasks. We experiment without the recurrence module (W/o recur), and {\name} without HuLM PT, and compare with \name. Bold indicates best in column and * indicates statistical significance $p < .05$ via permutation test w.r.t \name w/o recur. Results are reported in weighted F1.}
\label{tab:ablation} 
\end{table}

\noindent\textbf{Pre-training}
We assess the impact of pre-training by evaluating the downstream performance of a version of the \name model that has not been pre-trained on the HuLM task. Instead of using the weights from HuLM pre-training, we use \name with initialized weights as described in Section \ref{eval_lm}. Table \ref{tab:ablation} shows HuLM pre-training benefits -- pre-training adds substantial gain of 14.78 points and 4.84 points in weighted F1 for sentiment analysis and stance detection respectively.

\noindent\textbf{Recurrence}
We assess the importance of recurrent user state by first pre-training \name without its recurrent module and then fine-tuning it for the downstream tasks. We still use the same batching as described in Section \ref{ulm_task} but the information from a block no longer propagates to the next block in the forward pass, and backpropagation is still done on all blocks of a user together.
Without the recurrence module we see a drop of 1.21 points and 2.37 points in the weighted F1 measure for sentiment and stance respectively. Interestingly, \name outperforms \name without recurrence, consistent with the idea that models benefit from user history on tasks that involve a user.

%% file: conclusions.tex
\section{Conclusions}
Language is deeply human. Yet, language models in wide-spread use today lack a notion of the human that generates the language. Motivated by other advances in human-centered language processing and psychological theory that suggest language is moderated by human states, we introduced \textit{human language modeling}. 
\hulm extends LMs with the notion of a user and their states via their previous messages. 
In this first step toward large human language models, we developed a human-aware transformer (\name) that uses a recurrence mechanism to model user states and show that
pre-training this transformer on the human language modeling task yields significant gains in both generation and fine-tuning for multiple downstream document- and user-level tasks. 

Overall, state-of-the-art results with \name, a model neither trained on substantially larger data nor adding many parameters, suggests progress for transformers not based on massive increases in data or parameters but on a task grounded in language's ``natural'' generators, people. 

\section{Ethical Considerations}

While the multi-level human-document-word structure within \hulm can enable bias correcting and fairness techniques (discussed next), the ability to better model language in its human context also presents opportunities for unintended harms or nefarious exploitation.
For example, models that improve psychological assessment are not only useful for research and clinical applications, but could be used to target content for individuals without their awareness or consent.  
In the context of use for psychological research, such models may risk release of private research participant information if trained on private data without checks for exposure of identifying information. 
To negate this potential, we only release a version of \name that is without training on the consented-use private Facebook data until differential privacy standards can be verified.  

Unlike other human-centered approaches, \name is not directly fed user attributes as part of the pre-training thus the model parameters do not directly encode user attributes.

\hulm aims to join a growing body of work to make AI more human-centered, and thus more applicable for interdisciplinary study of the human condition as well as leading to new clinical tools for psychological health. 
At this point, our models are not intended to be used in practice for mental health care nor labeling of individuals publicly with personality or age scores.
While modeling the human state presents opportunities for reducing AI bias, prior to clinical or applied use, such models should be evaluated for failure modes such as error across target populations for error or outcome disparities~\cite{shah2020predictive}.
All user-level tasks presented here were reviewed and approved or exempted by an academic institutional review board (IRB). 

\section{Acknowledgments}
This work was supported by DARPA via Young Faculty Award grant \#W911NF-20-1-0306 to Stony Brook University; the conclusions and opinions
expressed are attributable only to the authors and should not be construed as
those of DARPA or the U.S. Department of Defense.
This work was also supported in part by NIH R01 AA028032-01 and by the National Science Foundation under the grant IIS-1815358.

%% file: appendix.tex
\appendix

\section{Appendix}
\label{sec:appendix}

\subsection{Pre-training}
\label{sec:app_pt}

\textbf{Twitter Data Collection} As mentioned, in section \ref{ulm_task}, we use a combination of data from both Twitter and Facebook data sources. However, since the main Twitter corpus~\cite{giorgi-etal-2018-remarkable} only spans the years 2009 - 2015, we wanted to supplement our total corpus with newer language data. Generally, we follow the same procedures for data collection as introduced for the 2009 - 2015 years. Thus, we started with a 1\% random sample of \textit{publicly available} tweets that can be mapped to US counties. On top of this we also applied the following filters: (1) Removal of non-English tweets, (2) Removal of users who did not tweet at least 3 times a week, (3) Removal of any duplicates among the collected data, and (4) Removal of any tweets containing URLs. We will be including this additional data as part of the CTLB project\footnote{\href{https://github.com/wwbp/county_tweet_lexical_bank}{https://github.com/wwbp/county\_tweet\_lexical\_bank}}.  

\begin{figure*}
    \centering
    \includegraphics[width=1.0\linewidth]{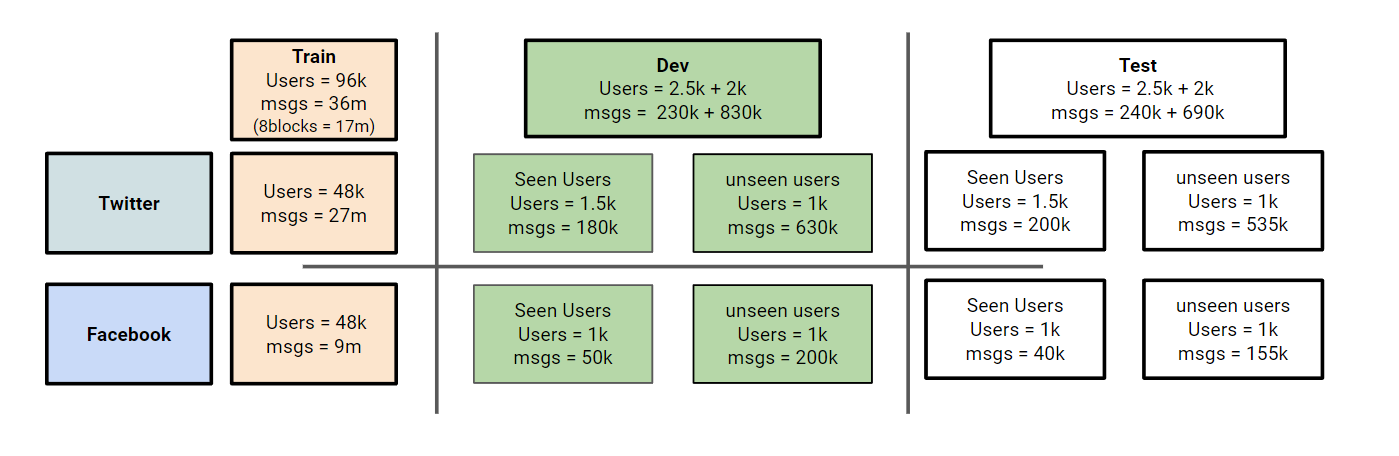}
    \caption{Structure of our pre-training dataset visually showing the data source(FB vs Twt), training/development/testing splits, and seen/unseen users for training and testing. Our dataset totals 100,000 users and approximately 37 million messages. Due to GPU memory restrictions, we limited training to 8 blocks of history which brought our train dataset size to 17 million messages. Dev and Test sets were not limited during evaluation.}
    \label{fig:pt_data}
\end{figure*}

\textbf{Data Size and Splits} We sample evenly between Facebook and Twitter at the user-level to collect 50,000 from each and apply the same minimum language use requirement of 1,000 words spanning 50 messages. We show the details of the splits across training/development/testing as well as seen/unseen user categories in figure \ref{fig:pt_data}. We keep 4,000 users for development and testing, 2k for each split, that are not at all present in the training portion. For users that we do train on, we select 4,500 to keep 20\% of their messages for development and testing sets.  

\subsection{Perplexity on Seen versus Unseen Users}
\label{eval:seen_ppl}
\paragraph{Benefit of Seen users.} By default, our experiments are run under an `unseen user' condition where by the test corpus contains users that were not in \name's training corpus. However, one could argue that this is an unnecessary impairment since further training the human language model doesn't require labels and can often be run on test data. 
We compare the effect of having seen users during \name training by additionally calculating perplexity on test sets with seen users. To make it a fair comparison, since we found our ``seen user'' corpus was more difficult (perplexity on seen users test set was higher than unseen users test set for \gptlmft as well), we use an adjusted perplexity, defined as the ratio of the model’s perplexity divided by a non-\hulm upper-bound perplexity on the same test set (\gptlmft), normalizing by the difficulty of the test set. As shown in Table \ref{tab:adj-ppl}, we find a small but significant benefit to having seen the users during training.

\begin{table}[t]
\begin{center}
\begin{tabular}{|l|p{.75cm}p{1.1cm}|p{.75cm}p{1.1cm}|}
\hline \bf  & \multicolumn{2}{|c|}{Unseen users}& \multicolumn{2}{|c|}{Seen users}\\
\bf Model & \small\it{ppl} & \small\it{adj-ppl}  & \small\it{ppl}& \small\it{adj-ppl}\\ 
\hline
\gptlmft & 48.5 & 1.00 &  53.7  & 1.00  \\
\name &  27.5 & \ 0.57* & 27.6 & \bf{0.51*} \\

\hline
\end{tabular}
\end{center}
\caption{\label{tab:adj-ppl} Evaluation of benefit of having seen the users during \name training. We use adjusted perplexity (\textit{adj-ppl}): the ratio of the perplexity to the upper-bound from not using \name during training (i.e.\gptlmft) on the same test set -- lower implies better performance when normalized by difficulty of the test set. Seen users test set is the set with the messages from the users also available in the train set, while unseen users test set does not have users common with the train set and is the same as the test set in Table \ref{tab:ppl}. Seen users test set is harder for both models. However, normalizing the scores show \name to have better performance over seen users test set. Bold font indicates best in column and * indicates statistical significance $p < .05$ via permutation test.}
\end{table}

\subsection{Experimental Settings}
\label{expmt_settings}
We use Open AI's pre-trained GPT-2 base model from \citet{radford2019language}  made available by the Hugging Face library from \citet{wolf2019huggingface} (transformers version 4.5.1) as our base model. We also make use of Hugging Face's code base to implement HuLM. Our training procedure involves all the default training hyperparameters from Hugging Face's GPT2 config except learning rate and the other specific hyperparams mentioned in the paper. We run a learning rate search sweep on a sampled dataset, for both {\name} and \gptlmft, using the Optuna framework from \citet{akiba2019optuna}: 1) in a range of 5e-6 to 5e-4, with 3 trials each of 5 epochs for pre-training, 2) in a range of 5e-6 to 5e-4, with 10 trials each of 15 epochs for fine-tuning stance detection, and 3) in a range of 1e-7 to 1e-5, with 5 trials each of 15 epochs for fine-tuning sentiment analysis. We also setup an early stopping criteria for the downstream task trials, such that we continue the epoch runs till we hit an increase in loss for 3 consecutive runs, and pick the model with the best F1 score. We couldn't run a similar sweep for user-level tasks due to compute time limits so we try a couple learning rates from document-level tasks but found the same learning rate that we use for pre-training to be better.  Many of the experimental/hyperparameters (batch sizes, window sequence sizes and cappings) settings mentioned throughout this work including the number of trials and the number of epochs vary because of computational limitations based on data size and training time. \\ 
All pre-training runs are trained on 2 Tesla V100 GPUs of ~32GB. Training \name takes approx 16 hours for 1 epoch (with train data consisting of 8 blocks (each of 1024 tokens) of 96000 users). Fine-tuning tasks run on a mix of Tesla V100, Quadro RTX 8000, and A100 GPUs based on compute availability. All batch sizes mentioned are per GPU.
\paragraph{Pre-training Settings} We use 2.4447e-4 as the learning rate for training \name, with 1 user train batch size, 15 users eval batch size and early stopping patience set to 3. For \gptlmft, we use the default settings from \citet{wolf2019huggingface} with train and eval batch size set to 60 and early stopping patience set to 3.
\paragraph{Document-level Fine-tuning Settings} We fine-tune \name for document-level tasks on their respective training data with an input instance capped to 8 blocks of 1024 tokens each, and no capping during evaluation. We train for 15 epochs using train and dev sets - along with history where available - with 1 user train batch size, 20 users eval batch size and early stopping patience set to 6. All models converge within 5 epochs except one stance target - feminism. \gptlmft is fine-tuned with the same data - but not history - using the same settings except a different learning rate (from the hyperparameter sweep mentioned above), train and eval batch size of 60, and max tokens per message set to 200 (consistent with pre-training).
\paragraph{User-level Fine-tuning Settings} We fine-tune \name for user-level tasks with an input instance capped to 4 blocks of 1024 tokens each, and evaluation data capped to 63 blocks (to allow for dev set evaluation due to compute limitations). For fine-tuning {\name}, we use 4 user train batch size and 20 eval batch size with early stopping patience set to 3. We layer norm the user-states (hidden states of the user state vector) from {\name}, and linearly transform (to embedding dimensions) before averaging the user-states to make the user's age estimation. We train for 30 epochs with warmup steps equivalent to 10 epochs, and a weight decay set to 0.01. We find that for the task of Age estimation the model converges at epoch 21, however for Personality Assessment we find a simple classification linear layer to show better performance (with a convergence seen at epoch 28 when run for 35 epochs). In case of \gptlmft we with the same data (split into into individual messages capped to 200 tokens per message as in pre-training), for 15 epochs (much higher training time as compared to {\name}) with train and eval batch size set to 400, and early stopping patience set to 3.
\paragraph{MeLT -- Sentiment Fine-tuning Settings} To apply MeLT \cite{matero-etal-2021-melt-message} to the sentiment task we use use optuna~\cite{akiba2019optuna} to search both learning rate and weight decay parameters using a search space between 6e-6 and 3e-3 and between 1 and 1e-4 respectively. We keep the same architecture as described in the original MeLT paper, however we make 1 change during fine-tuning and that is the message-vector representation from MeLT is concatenated with the average of the observed tokens for the labeled message to include both local and global context into the fine-tuning layers. 
\paragraph{No Historical Context Fine-tuning Settings} We run a hyperparameter sweep using Optuna \cite{akiba2019optuna} for all models for learning rate (using search space between 5e-6 to 5e-4) and weight decay(using search space between 0.0 and 1.0) with early stopping patience set to 6. We do this for 15 and 10 trials for Stance and Sentiment models respectively, and pick the hyperparameters value for the best model in the same way as described in the Experimental Settings (\ref{expmt_settings} section above. We use these values to fine-tune the models for 15 epochs and get the weighted F1 results.

\begin{table}[t]
\begin{center}
\begin{tabular}{|p{3.6cm}|l|l|}
\hline \bf{Model} &
\it{Age ($r$)} & 
\it{OPE ($r_{dis}$)} \\ 
 
\hline
\name (Full test set) & 0.868 & 0.619 \\
\name (Filtered test set) & 0.872 & 0.635 \\

\hline  
\end{tabular}
\caption{\label{tab:user_ft_filtered} \name's performance on user level downstream tasks: Age and Openness (OPE), on full test sets (5000 users and 1943 users respectively for Age and OPE) from \citet{Park2015AutomaticPA} and \citet{lynn-etal-2020-hierarchical}, as well as on the resulting test set (4438 users and 1745 users respectively for Age and OPE) after filtering the dataset for English language with users having a minimum of 50 posts and 1000 words (as we do for our pre-training data).
}
\end{center}
\end{table}